\documentclass[runningheads]{llncs}
\usepackage{graphicx}
\usepackage[sorting = none, backend = bibtex, style= nature]{biblatex}
\usepackage{hyperref}

\begin{document}
%
\title{A view on model misspecification in uncertainty quantification}

%
\author{Yuko Kato\inst{1} \and
David M.J. Tax\inst{1}\and
Marco Loog\inst{1,2}}
\authorrunning{Y.~Kato et al.}
%
\institute{Delft University of Technology \and University of Copenhagen}
\maketitle     
\renewcommand{\footnoterule}{%
  \kern -0.4pt
  \hrule width \textwidth height 0.4pt
  \kern 1.4pt
}
\let\thefootnote\relax\footnote{An initial version of the current work has been accepted to be presented at BNAIC/BeNeLearn 2022, to which it was submitted August 27, 2022.}
%

\begin{abstract}
Estimating uncertainty of machine learning models is essential to assess the quality of the predictions that these models provide. However, there are several factors that influence the quality of uncertainty estimates, one of which is the amount of model misspecification. Model misspecification always exists as models are mere simplifications or approximations to reality. The question arises whether the estimated uncertainty under model misspecification is reliable or not. In this paper, we argue that model misspecification should receive more attention, by providing thought experiments and contextualizing these with relevant literature. 


\keywords{Uncertainty quantification \and Model misspecification \and Epistemic and Aleatoric uncertainty}
\end{abstract}
\section{Introduction}
In fields such as biology, chemistry, engineering, and medicine \cite
{Hie2020-st, Vishwakarma2021-gb, Begoli2019-tl, Michelmore2018-in}, having an accurate estimate of prediction uncertainty is of great importance to guarantee safety and prevent unnecessary costs. In this regard, uncertainty quantification (UQ) is a vital step in order to safely apply Machine Learning (ML) models to real world situations involving risk. It is well-known, however, that these ML models are generally poor at quantifying these uncertainties  \cite{Lakshminarayanan2017-ld, Maddox2019-rs}. 

Generally, depending on the exact source, the type of uncertainty can be categorized as being epistemic or aleatoric \cite{Kiureghian2009-hk, Abdar2021-fu}. Epistemic uncertainty refers to the uncertainty of the model and arises due to lack of data used to train the model and lack of domain knowledge. This type of uncertainty is considered to be reducible by an appropriate selection of the model and increasing data size. On the contrary, aleatoric uncertainty is a consequence of the random nature of data and is therefore considered to be irreducible \cite{Hullermeier2021-uk}.  

Although total uncertainty (the combination of epistemic and aleatoric uncertainty) can be estimated using different methods, some articles have focused on the separate estimation of aleatoric uncertainty and epistemic uncertainty \cite{Depeweg2017-pz, Kendall2017-uy, Senge2014-uw, Prado2019-vs, Gustafsson2020-op}. Given the fact that only epistemic uncertainty is reducible, separation into the two types of uncertainty can help to guide model development \cite{Senge2014-uw}. For example, during active learning, Nguyen and colleagues \cite{Nguyen2022-xa} concluded that quantifying epistemic uncertainty separately has the potential to provide useful information to learners of the model and can potentially improve the performance of the learning process. 

Notwithstanding, due to model misspecification, a question remains whether the estimated aleatoric and epistemic uncertainty are reliable or not  \cite{Cervera2021-jq,Lv2014-fi}. It is important to realize that model misspecification always exists without any exceptions \cite{Aydogan2020-te}. As
there is no uniquely prescribed way to deal with model misspecification, it has been treated differently among researchers. Some do not mention it at all in their papers \cite{Kendall2017-uy, Gustafsson2020-op}, while others consider it to be part of epistemic uncertainty \cite{Lahlou2021-tl}. These variable interpretations of model misspecification make it difficult to compare the different estimated epistemic uncertainties.

In this paper, we argue that model misspecification should receive more attention. We start by defining model misspecification and propose three possible ways to see model misspecification in relation to epistemic uncertainty. The paper proceeds with a brief investigation of how model misspecification is recognized among researchers and how they relate to our proposed views. In addition, we assess the possible consequences of mistreating model misspecification qualitatively and conclude with a brief discussion.

\section{Model misspecification} 

\begin{figure}
\centering
\includegraphics[scale = 0.85]{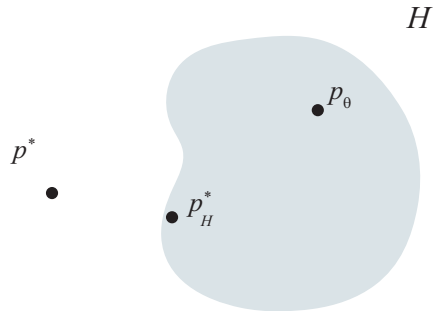}
\caption{Model misspecification: The truth $p^*$ is not included in the hypothesis space $H$ (shaded blue area) due to model misspecification.}  \label{fig1}
\end{figure}

\subsection{Definition of model misspecification}
Whether the ML-method is applied in a regression or classification setting,  the first step is to choose a model $p_{\theta}$ parameterized by $\theta$. Let $H = \{p_{\theta}:\theta \in \Theta\}$ denote the hypothesis space defined by the chosen model class and the associated parameters $\theta$ in the set $\Theta$. We denote the truth $p^*$ and the best possible model in $H$ by $p^*_{H}$. Now, model misspecification exists when the best model $p^*_{H}$ differs from the truth $p^*$. In other words, model misspecification exists if   $p^* \notin H$. This is illustrated in Fig.~\ref{fig1}.

\subsection{Example of model misspecification: regression}
\label{example}
\begin{figure}
\centering
\includegraphics[scale = 0.65]{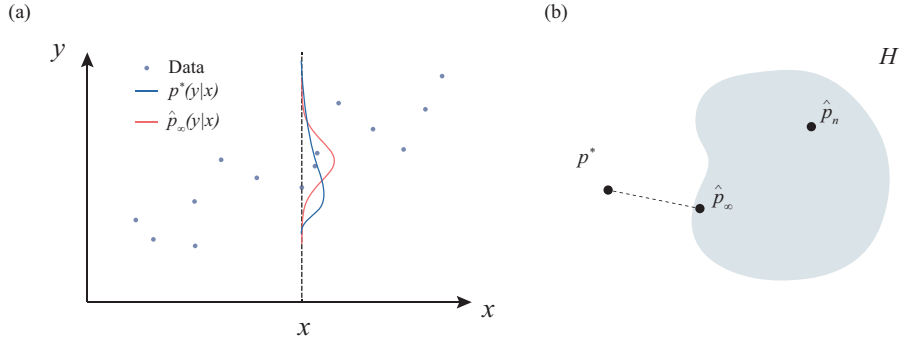}
\caption{Model misspecification in a regression example: (a) the true data distribution $p^*(y |x)$ (blue line) at a test point $x$ and the estimated data distribution $\hat{p}_\infty (y|x)$ (red line) which is trained using an infinite amount of data at a test point $x$ (b) $\hat{p}_\infty$ will never reach the truth $p^*$ -- even when optimizing for a proper loss, due to model misspecification.}  \label{fig2}
\end{figure}

Let us consider a simple regression problem using a dataset of $n$ observations $\mathcal{D} = (x_i, y_i)$ with $i = 1,...,n$ and additive noise. Let us assume, in addition, that the  true additive noise follows a heavy-tailed  distribution (see Fig.~\ref{fig2}a). The typical  aim is to approximate the unknown underlying true data distribution $p^*(y |x)$ at a new test point  $x$. Now, a typical model choice $\hat{p}_n(y |x)$ is to assume the additive noise to be Gaussian. This means that the assumption on the noise distribution is misspecified. Since $p^*(y |x)$ is not in the hypothesis space $H$ due to model misspecification (i.e., the assumed noise does not match the true one), we will never be able to reach the true distribution $p^*(y |x)$ even if we use an infinite amount of data to train the model $\hat{p}_{\infty}(y |x)$ (See Fig.~\ref{fig2}b). 

\subsection{Contextualizing model misspecification}

\begin{figure}
\centering
\includegraphics[scale = 0.6]{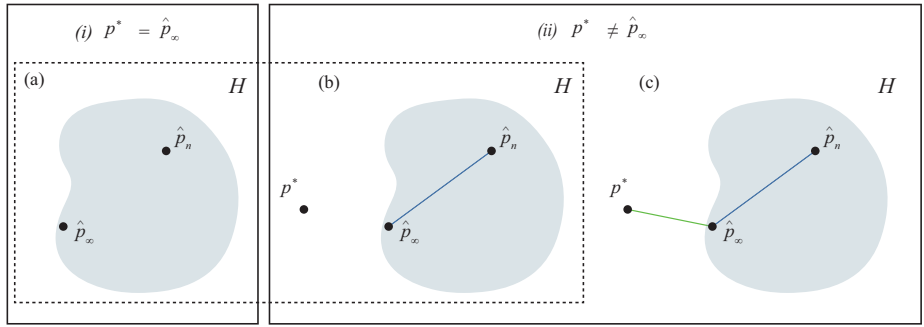}
\caption{Three views on model misspecification: In (a) and (b), model misspecification is ignored and epistemic uncertainty (blue line) is defined as a difference between $\hat{p}_{\infty}(y|x)$ and $\hat{p}_n$. In (c), model misspecification is treated as a part of epistemic uncertainty (green line).} \label{fig3}
\end{figure}

 When looking at model misspecification, a key uncertainty we consider is epistemic uncertainty which is related to the model choice \cite{Aydogan2020-te}. Fig.~\ref{fig3} shows three perspectives on model misspecification and its relation to epistemic uncertainty for regression setting. 

The first two scenarios (Fig.~\ref{fig3}a and Fig.~\ref{fig3}b) illustrate the most common perspective where model misspecification is ignored. Ignoring it could be acceptable if the hypothesis space was sufficiently large and, therefore, model misspecification does not exist in principle (Fig.~\ref{fig3}a). Even though most practitioners disregard model misspecification as illustrated in this Fig.~\ref{fig3}a, in reality this is rarely the case. The most common circumstance in the literature is shown in Fig.~\ref{fig3}b, where model misspecification exists despite being unintentionally ignored. As a result, for both scenarios, epistemic uncertainty does not contain model misspecification. In the third scenario (Fig.~\ref{fig3}c), model misspecification is explicitly included as part of epistemic uncertainty. In this case, epistemic uncertainty consists of two parts: model misspecification (green line in Fig.~\ref{fig3}c) and approximation uncertainty (blue line in Fig.~\ref{fig3}c) as H{\"u}llermeier and Waegeman \cite{Hullermeier2021-uk} define it. In this scenario, model misspecification directly influences epistemic uncertainty estimates \cite{Xu2022-jm}.

\section{Effect of model misspecification on uncertainty estimates}
We cover, in brief, the connection between model misspecification and epistemic uncertainty in different ML-methods and discuss possible consequences of model misspecification on uncertainty estimation and decision making.

\subsection{Model misspecification and uncertainty estimates}
Bayesian ML-methods are widely used for UQ \cite{Tipping2004-sr}. Specifically, Bayesian neural networks (BNNs) are receiving a lot of attention due to their potential to model both epistemic and aleatoric uncertainty \cite{Kendall2017-uy, Scalia2020-sa, Depeweg2017-pz, Gustafsson2020-op}. The definition of uncertainty varies among researchers and BNN articles rarely include the effect of model misspecification. For instance, Gustafsson and colleagues \cite{Gustafsson2020-op} explain that epistemic uncertainty is  uncertainty in the deep neural network model parameters, which ignores the fact that the model is not necessarily correctly specified. This definition of epistemic uncertainty is very common in the literature \cite{Depeweg2017-pz, Hullermeier2021-uk, Mobiny2021-gc, Gal2016-vk}. 

Without loss of generality, consider once again the regression setting from Subsection \ref{example}. In BNNs, the total uncertainty is modelled by the posterior $p(\theta|\mathcal{D})$, where $\mathcal{D}$ symbolizes data. At test time, predictions are made via the posterior predictive distribution (PPD):

\begin{equation}
p(\mathbf{y} |~ \mathbf{x},\mathcal{D})= \int p(\mathbf{y} |~ \mathbf{x},\theta)p(\theta|\mathcal{D})d\theta.    
\end{equation}

In \cite{Gustafsson2020-op}, both epistemic and aleatoric uncertainties are obtained assuming a Gaussian distribution with mean $\hat{\mu}$  and variance $\hat{\sigma^2}$ on PPD:
\begin{equation}
p(\mathbf{y} |~ \mathbf{x},\mathcal{D}) \approx N(\mathbf{y}; \hat{\mu} (\mathbf{x}), \hat{\sigma}^2 (\mathbf{x})).    
\end{equation}
It is safe to say that that no true distribution is exactly Gaussian. The possible effects on UQ of such assumption are neither quantified nor discussed in this work; not as part of epistemic uncertainty itself, nor as a separate uncertainty. In the literature, the (implicit) assumption of a PPD with a Gaussian distribution seems to be made more generally when creating Bayesian ML-models \cite{Depeweg2017-pz, Hullermeier2021-uk, Mobiny2021-gc, Gal2016-vk}.

Another popular approach for UQ is the use of ensemble methods, such as Monte
Carlo dropout (MC-dropout) \cite{Gal2016-vk}, where predictive uncertainty is estimated using
Dropout \cite{Gal2016-vk} at test time, and Deep ensembles \cite{Lakshminarayanan2017-ld} which rely on retraining
the same network many times with different weight initializations. Both methods can be considered a simple alternative to Bayesian methods. It is known that ensemble methods can
provide an estimation of the (epistemic) uncertainty of a prediction \cite{Hullermeier2021-uk}, meaning
that the variance of the predictions can be used to estimate epistemic uncertainty.
By increasing the number of ensemble members, improved estimation
of epistemic uncertainty is possible \cite{Hullermeier2021-uk, Charpentier2022-dp, Caldeira2020-or}.

Liu and colleagues pointed out two issues when performing uncertainty estimation (both epistemic and aleatoric) using ensemble methods \cite{Liu2019-yx}. Similar to Bayesian methods, currently existing ensemble methods typically assume that the ground-truth data distribution $p(y|x)$ follows a Gaussian distribution \cite{Lakshminarayanan2017-ld}. Furthermore, ensemble methods perform uncertainty estimation using base models of the same class, meaning in the same hypothesis space $H$ \cite{Lakshminarayanan2017-ld}. The consequence is that the creation of these models (that are all potentially misspecified) might result in a hypothesis space that still does not include the true distribution. Therefore epistemic uncertainty estimates from ensemble methods do not include model misspecification. 




\subsection{Consequences of ignoring model misspecification}
The central question that remains is what are the consequences of not taking model misspecification into account? 

\begin{figure}
\centering
\includegraphics[scale = 0.6]{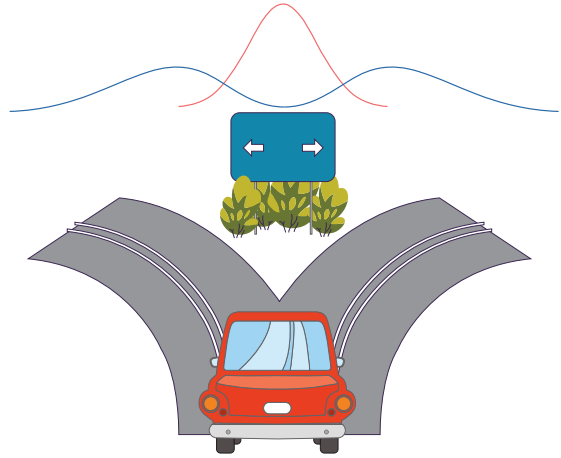}
\caption{A real-world example of model misspecification: In a decision making process at a split road, assuming a Gaussian distribution on the output distribution (red line) can cause a serious accident} \label{fig4}
\end{figure}

If we do not include model misspecification into epistemic uncertainty, we can only trust the estimated epistemic uncertainty when the model is correctly specified. As a result, we may significantly underestimate total uncertainty, which can be problematic in risk-involving tasks. Therefore, ignoring model misspecification usually leads to the scenario in Fig.~\ref{fig3}b which can overestimate or underestimate the total uncertainty \cite{Cervera2021-jq}. Specifically for Bayesian methods, when the distributional assumption on PPD is not correctly specified, the probability of the true distribution lying outside hypothesis space increases \cite{Masegosa_undated-sg, Jewson2018-oy, Cervera2021-jq, Valdenegro-Toro2022-dj, Hansen_undated-nc, Ramamoorthi2015-ps}. The consequences of ignoring model misspecification in real-world tasks are illustrated in Fig.~\ref{fig4}, which is based on an example from \cite{Cervera2021-jq}.

In Fig.~\ref{fig4}, a lane splits in two directions, i.e., we have two outputs. In order to capture the multimodality, our output distribution has to follow a bi-modal distribution. However, by assuming a Gaussian distribution on the output  distribution instead, the model predicts the mean of the two possible outcomes and the car goes in between lanes \cite{CerreiaVioglio2020-rz}. Although this can lead to a fatal accident, this possibility cannot be captured by estimating epistemic uncertainty with such model.  


\section{Discussion}
In this paper, we highlighted the importance of considering model misspecification when performing UQ. Reviewing the literature, it should be noted that model misspecification is often not explicitly described or strong assumptions are imposed on the underlying data distributions. This can lead, in turn, to models that provide unreliable uncertainty estimates. These findings raise an important question: how should we handle model misspecification? As for the definition of different uncertainty types (i.e., aleatoric and epistemic
uncertainty), the question remains whether epistemic uncertainty should contain model misspecification or not. These issues are discussed in the following Subsection \ref{how to handle mm} and Subsection \ref{def of epi and ale}.

\subsection{How to handle model misspecification
}
\label{how to handle mm}
It is not possible to entirely avoid model misspecification when modeling real-world phenomena, mainly due to the imposed model assumptions. This means that we have to consider the impact of model misspecification and to explore ways to deal with it in various scenarios.   

\begin{figure}
\centering
\includegraphics[scale = 0.65]{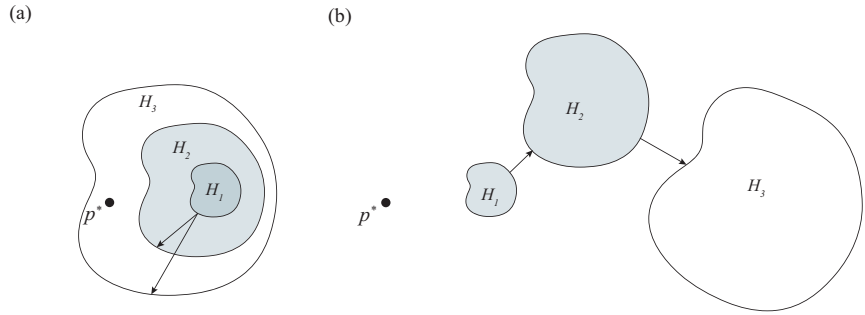}
\caption{Reducing model misspecification: By changing the initial hypothesis space $H_1$, model misspecification is either (a) reduced by including the truth $p^*$ in the new hypothesis space $H_3$ or (b) increased} \label{fig6}
\end{figure}

In principle, model misspecification can be reduced by expanding the initial hypothesis space $H_1$ (i.e., changing the associated model) (Fig.~\ref{fig6}a). As a result, the impact of model misspecification can be reduced subsequently. Expanding the hypothesis space can be done, for instance, by easing the imposed assumptions. In the most favorable situation, the expanded hypothesis space $H_3$ includes the truth $p^*$. However, there is a possibility that we increase model misspecification when changing hypothesis space (Fig.~\ref{fig6}b). As a result, we would never be able to reach the $p^*$ in $H_3$. This situation can be avoided by assigning a hierarchy to the expanded hypotheses spaces as follows, $H_1 \subset H_2 \subset H_3$. Under this assumption, structural risk minimization (SRM) \cite{Guyon1991-nm} , which is strongly universally consistent \cite{Lugosi2002-cn}, can be arbitrary close to the $p^*$ in $H_3$. SRM uses the size of hypothesis space as a variable and tries to minimize the guaranteed risk over each hypothesis space  \cite{Corani2007-nc, Zhang2010-nj}. This can reduce model misspecification. 

Therefore, in either way (Fig.~\ref{fig6}a or Fig.~\ref{fig6}b), the impact of model misspecification can be minimized. This fact is important since it is not always clear if we change our hypothesis space inclusively (Fig.~\ref{fig6}a) or exclusively (Fig.~\ref{fig6}b). 

Additionally, it is theoretically possible to reduce model misspecification completely according to the universal approximation theorem \cite{Hornik1989-ej, Cybenko1989-uo}. The theorem guarantees a neural network to represent any function (e.g., input-output relationship in regression) generically on its associated function space  \cite{Kratsios2021-sa}. However, the theorem cannot be applicable to most practical situations due to the following reasons. Firstly, the theorem does not guarantee the network to learn the model \cite{Goodfellow2016-fz}. Therefore, the chosen network can overfit to the training data, resulting in a poor generalization. Secondly, the  network can require an exponential number of hidden units, which are often not desirable in practical situations \cite{Goodfellow2016-fz}. These concerns lead us to the next question; What are the consequences when we attempt to reduce model misspecification during UQ?

\begin{figure}
\centering
\includegraphics[scale = 0.8]{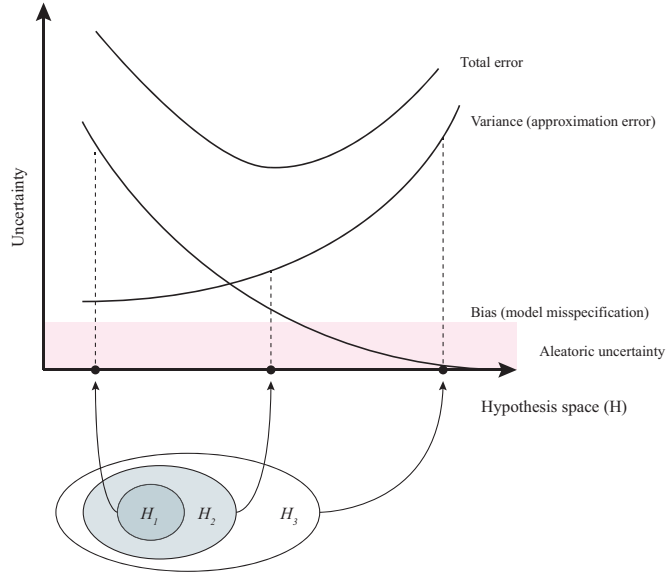}
\caption{Intuition for bias (model misspecification) variance (approximation error) trade-off with respect to the size of hypothesis space. Red shaded area represents aleatoric uncertainty.} \label{fig7}
\end{figure}

In order to consider the question, let us go back to the scenario where model misspecification is considered to be part of epistemic uncertainty (Fig.~\ref{fig3}c). Lahlou and colleagues \cite{Lahlou2021-tl} state that model misspecification can be considered as a form of bias while approximation uncertainty can be a form of variance. With the use of this concept, we can think about a bias-variance tradeoff with respect to the size of hypothesis space, which is illustrated in Fig.~\ref{fig7}. 


In Fig.~\ref{fig7}, we can see the effect of model misspecification in relation to total uncertainty. Assume that we have an infinite amount of data, then every hypothesis space shares the same amount of aleatoric uncertainty (red shaded area in Fig.~\ref{fig7}). Depending on the model choice, the size of hypothesis space can vary. If we choose the initial hypothesis space $H_1$, there is a high possibility that $p^*$ is not included in the hypothesis space, potentially increasing model misspecification and therefore bias of the model.
On the contrary, if we choose an expanded hypothesis space $H_3$, model misspecification can be decreased due to the fact there is a higher possibility that $p^*$ lies in $H_3$. However, at the same time, it will be harder to find an optimal model in an expanded hypothesis space. Therefore, the variance which represents approximation uncertainty increases when hypothesis space expands. This means that there is trade-off between bias (model misspecification) and variance (approximation error). In this example, hypothesis space $H_2$ has an optimal size. In this case, $H_2$ results in the best trade-off between model misspecification and its size.  Using this concept, we can choose the best possible model and the associated hypothesis space, even if we do not know $p^*$.     

However, whether the approximation error and model misspecification can be estimated in practical scenarios is not entirely clear yet. This means that how to handle model misspecification still remains an open question.  


\subsection{Conflicting views on uncertainty definitions}
\label{def of epi and ale}
\begin{figure}
\centering
\includegraphics[scale = 0.7]{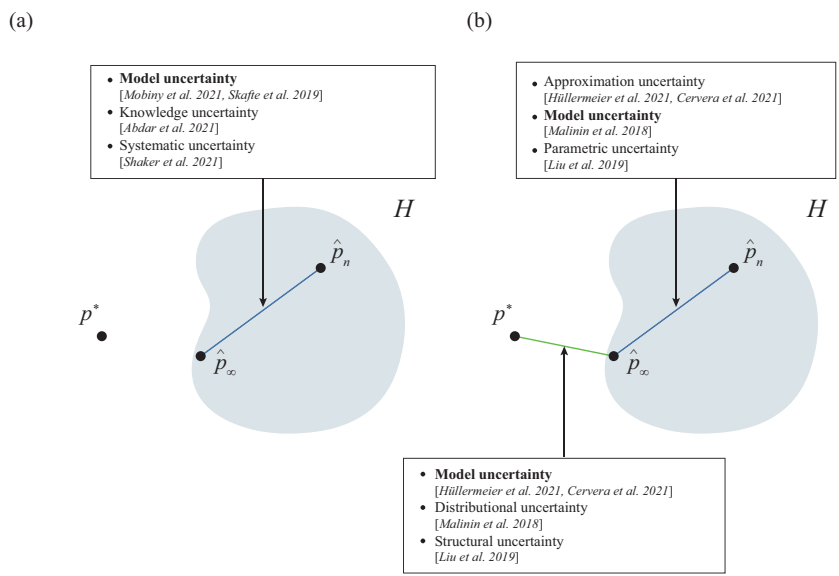}
\caption{Example of variability in naming of epistemic uncertainty: (a) Model misspecification is ignored. (b) Model misspecification is considered as part of epistemic uncertainty (a green line). Model uncertainty represents different concepts depending on literature} \label{fig5}
\end{figure}

There are conflicting views and significant terminology diversity about the different types of uncertainty in the literature. In particular, model misspecification has been included as part of epistemic uncertainty by some authors while others do not \cite{Masegosa_undated-sg, Liu2019-yx}. This is problematic since it prevents a direct comparison regarding the performance of different methods. A uniform view on how to treat model misspecification (either being
part of epistemic uncertainty or as a separate uncertainty type) should be made
to make such comparison possible. Furthermore, it is not very clear which type of uncertainty is estimated in some situations. For example, Valdenegro-Toro and colleagues \cite{Valdenegro-Toro2022-dj} claim that there is a clear connection between epistemic and aleatoric uncertainty. This would mean that underestimating or overestimating epistemic uncertainty can have an effect on the quality of aleatoric uncertainty. Therefore, a true possibility for such a separation can be doubted in this case. Additionally, there is another issue related to the definition of uncertainty, which is inconsistent naming of uncertainties. This variability in naming of epistemic uncertainty is illustrated in Fig.~\ref{fig5}. Note that this figure does not reflect an exhaustive literature search, so there is even more diversity in terminology than what is illustrated there.


Epistemic uncertainty is called as, for instance, model uncertainty \cite{Mobiny2021-gc, Skafte2019-ib}, knowledge uncertainty \cite{Abdar2021-fu}, systematic uncertainty \cite{Shaker2021-os}  (Fig.~\ref{fig5}a). In other work, model uncertainty is considered to be part of epistemic uncertainty (Fig.~\ref{fig5}b), representing model misspecification \cite{Cervera2021-jq, Hullermeier2021-uk}. Malinin and colleagues  \cite{Malinin2018-kj} call model misspecification as distributional uncertainty. This inconsistency can be confusing and leading to wrong interpretation of published work. 

\section{Conclusion}
There are a number of concerns when it comes to model misspecification in relation to uncertainty estimation that can considerably influence the accuracy by which uncertainty estimation can be performed. No general definition of both epistemic and aleatoric uncertainty currently exists. Additionally, researchers treat model misspecification differently. Since model misspecification influences the reliability of uncertainty estimates, we would argue that model misspecification, and the ways to measure and control it, should receive more attention in the current UQ research. Furthermore, we propose directions for future work to minimize model misspecification, for instance, by looking into bias-variance trade-offs.

\renewcommand{\bibfont}{\normalfont\small}
\printbibliography
\end{document}